\documentclass{article}
\usepackage{ijcai24}

\usepackage{times}
\usepackage{soul}
\usepackage{url}
\usepackage[hidelinks]{hyperref}
\usepackage[utf8]{inputenc}
\usepackage[small]{caption}
\usepackage{graphicx}
\usepackage{amsmath}
\usepackage{amsthm}
\usepackage{booktabs}
\usepackage{algorithm}
\usepackage{algorithmic}
\usepackage[switch]{lineno}
\usepackage{amsfonts}
\usepackage{color}
\usepackage{comment}
\usepackage{makecell}
\usepackage{subfigure}
\usepackage{pifont}
\usepackage{multirow}
\usepackage{bm}
\usepackage{colortbl}
\usepackage[table]{xcolor}

\usepackage{stfloats}
\newcommand{\cmark}{\ding{51}}%
\newcommand{\xmark}{\ding{55}}%

\definecolor{lightgray}{rgb}{0.9,0.9,0.9}
\definecolor{lightgray1}{rgb}{1,0.9,0.9}

\title{Contrastive Learning Meets Pseudo-label-assisted Mixup Augmentation: A Comprehensive Graph Representation
Framework from Local to Global}

\author{
Jinlu Wang$^1$
\and
Yanfeng Sun$^{1}$\and
Jiapu Wang$^{1}$\and
Junbin Gao$^{2}$\and 
Shaofan Wang$^{1}$\and 
Jipeng Guo$^{3,*}$
\\
\affiliations
$^1$ School of Information Science and Technology, Beijing University of Technology, China\\
$^2$The University of Sydney Business School, The University of Sydney, Australia\\
$^3$College of Information Science and Technology, Beijing University of Chemical Technology, China
\emails
wangjinlu@emails.bjut.edu.cn, yfsun@bjut.edu.cn, jpwang@emails.bjut.edu.cn, \\junbin.gao@sydney.edu.au, wangshaofan@bjut.edu.cn, guojipeng@buct.edu.cn
}

\begin{document}

\maketitle

\begin{abstract}
Graph Neural Networks (GNNs) have demonstrated remarkable effectiveness in various graph representation learning tasks. However, most existing GNNs focus primarily on capturing local information through explicit graph convolution, often neglecting global message-passing. This limitation hinders the establishment of a collaborative interaction between global and local information, which is crucial for comprehensively understanding graph data. To address these challenges, we propose a novel framework called \textbf{Com}prehensive \textbf{G}raph \textbf{R}epresentation \textbf{L}earning (\textbf{ComGRL}). ComGRL integrates local information into global information to derive powerful representations. It achieves this by implicitly smoothing local information through flexible graph contrastive learning, ensuring reliable representations for subsequent global exploration. Then ComGRL transfers the locally derived representations to a multi-head self-attention module, enhancing their discriminative ability by uncovering diverse and rich global correlations.  
To further optimize local information dynamically under the self-supervision of pseudo-labels, ComGRL employs a triple sampling strategy to construct mixed node pairs and applies reliable Mixup augmentation across attributes and structure for local contrastive learning. This approach broadens the receptive field and facilitates coordination between local and global representation learning, enabling them to reinforce each other. Experimental results across six widely used graph datasets demonstrate that ComGRL achieves excellent performance in node classification tasks. The code could be available at \url{https://github.com/JinluWang1002/ComGRL}

\end{abstract}

\section{Introduction} 
Graph Neural Networks (GNNs) have demonstrated remarkable   performance across various graph learning tasks, such as node classification \cite{kipf2017semi,velickovic2017graph,yang2024}, node clustering \cite{wang2023overview,zhao2022adaptive}, and link prediction \cite{zhang2018link,cai2022line}. Due to their powerful representation capabilities, GNNs perform effectively even on complex graph structures. However, most existing GNNs approaches rely heavily on neighbor aggregation mechanism for local feature extraction. For example, the classic Graph Convolutional Network (GCN) \cite{kipf2017semi} updates the representation of each target node by averaging features from the node itself and its neighbors. GraphSAGE \cite{hamilton2017inductive} introduces a sampling method
to select neighbors for each node, significantly reducing   computational complexity. The Graph Attention Network (GAT) \cite{velickovic2017graph} employs an attention mechanism to assign dynamic weights to neighboring nodes during aggregation, reflecting their relative importance. Meanwhile, the Graph Isomorphism Network (GIN) \cite{xu2018powerful} adopts a simple architecture, aggregating information from neighboring nodes through a weighted Multi-Layer Perceptrons (MLPs).

Despite their remarkable effectiveness, most existing GNNs adopt shallow architectures, which mainly retain more local information. In fact, global contextual information can provide critical insights for representation learning, enhancing the robustness of node classification by capturing long-range dependencies among nodes. A straightforward approach to expanding the receptive field of a target node is to deepen these networks by stacking additional GNN layers. Unfortunately, this often leads to the over-smoothing problem, which dimishes the model's discriminative power \cite{wu2021comprehensive}.

To address these challenges, some researches try to integrate global and local information to enhance the final node representations. The distinction between these approaches often lies in their mechanisms for exploring global and local information and their strategies for fusing two. Some methods propagate more distant information by modifying the message-passing matrix. For instance, the Local-Global Adaptive
Graph Neural Network (LG-GNN) \cite{yu2024lg} constructs local feature similarity and long-range structural similarity, fusing them to provide comprehensive guidance for feature propagation. Similarly, the HOmophily-Guided
GCN (HOG-GCN) \cite{wang2022powerful} considers all neighbors and utilizes a learnable mask matrix to adaptively aggregate global information. Transformers,  
with their self-attention mechanism, excel in globally aggregating information \cite{Vaswani2017attention}. Building on this, TransGNN \cite{zhang2024transgnn} alternates between GNNs and Transformers to mutually leverage their respective strengths, achieving a balanced integration of local and global information. In addition, the Local and Global Disentanglement for Graph Convolutional Networks (LGD-GCN) \cite{guo2022learning} utilizes both local and global information to disentangle node representations in the latent space. This is achieved through a neighborhood routing mechanism and disentangled global graph encoding. 

Nevertheless, the integration of local and global information to obtain comprehensive graph node representation still faces significant challenges, mainly including three crucial aspects. (1) \textit{How to effectively extract local information?} Most existing methods rely on graph convolution with the original topological graph to aggregate local information, which is heavily influenced by the topological structure. In general, the topological graph is sparse, fixed, and noisy, potentially prohibiting
models from learning rich local information. (2) \textit{How to adaptively extract complete global information?} Global information expands the receptive fields of information aggregation by focusing on
more relevant nodes. Optimal and abundant global information should be adaptively captured from various aspects. (3) \textit{How can local and global information be sufficiently coupled and collaboratively interact within a unified framework?} Considering the unique strengths of both local and global information, it is necessary to achieve effective integration and mutual interaction in a cohesive framework.

Based on the considerations above, this paper proposes a \textbf{Com}prehensive \textbf{G}raph \textbf{R}epresentation \textbf{L}earning framework, named as \textbf{ComGRL}, which integrates the strengths of local-and-global information to mutually enhance their functionality. Specifically, \textbf{in local information modeling}, ComGRL utilizes MLPs as feature extractors and flexibly smooths node representations by leveraging graph contrastive learning, rather than relying on rigid GNN layers. To elegantly combine local information with global information and provide a more robust input for global correlation modeling, the local representation is transmitted to a multi-head self-attention module, serving as the foundation for global information aggregation. Hence, \textbf{in global information modeling}, ComGRL captures diverse global correlations to derive a comprehensive final representation. Finally, \textbf{to achieve  mutual interaction between local and global information}, we propose a pseudo-label-assisted Mixup augmentation module, which uniformly guides these two modules in a self-supervised manner. ComGRL utilizes a node mixup augmentation with a triple sampling strategy, combining pseudo-labels of global representation with external prior knowledge. This approach enables graph augmentations and rectifications (including attributes and structure) for local contrastive learning. Through its collaborative architecture, ComGRL is able to assist the integration and interaction between local and global information better.

We summarize the main contributions of our paper from the following three aspects:
\begin{itemize}
    \item We propose a comprehensive node representation learning method, \textbf{ComGRL}, in which local graph contrastive learning (LGCL) and global multi-head self-attention (GMSA) mutually collaborate within a unified framework. GMSA expands information propagation, while LGCL flexibly captures essential structural information for GMSA,  enhancing its expressive ability. 
    \item To achieve self-supervision and mutual reinforcement between LGCL and GMSA, we develop a Pseudo-label-assisted Mixup Augmentation (PMA) mechanism. This mechanism combines the relatively reliable pseudo-label information from GMSA with external priors to adaptively guide data augmentation for LGCL.
   \item Finally, we conduct semi-supervised node classification experiments on six widely used graph datasets, where ComGRL achieves excellent performance across all the datasets.
\end{itemize}

\section{Preliminaries}
\subsection{Notations}

Let $\mathcal G=(\mathcal V, \mathcal E, \mathbf X)$ denote an undirected graph with a node set $\mathcal V=\{v_1, \cdots, v_N\}$, an edge set $\mathcal E$, and node attribute features $\mathbf X = [\mathbf x_1, \mathbf x_2, \cdots, \mathbf x_N]^T \in \mathbb R^{N \times D}$, where $D$ is the dimension of the node features. The connection relationships of graph $\mathcal G$ can be mathematically described by the adjacency matrix $\mathbf A = [a_{ij}]_{i,j=1}^N \in \{0,1\}^{N \times N}$, and each element $ a_{ij}$ in $\mathbf A$ represents the connectivity relationships between nodes $v_i$ and $v_j$, i.e.,
\begin{equation}
 a_{ij}=
\begin{cases}
1 & \text{if} \ \ (v_i,v_j) \in \mathcal E;\\
0 & \text{otherwise}.
\end{cases}
\end{equation}
Given $\mathbf X$, $\mathbf A$, and the training set $\mathcal V_L\subset \mathcal{V}$ with true labels $\mathbf Y_L$, the node classification task aims to train a network $f(\cdot)$ by minimizing loss $\mathcal{L}(f(\mathbf X,\mathbf A ), \mathbf Y_L)$, then the  $f(\cdot)$ with classifier can accurately predict the labels $\mathbf Y_U$ for nodes in the test set. The discriminative node representation is crucial for excellent classifier. In the GNNs based node classification paradigm, neighbor information aggregation mechanisms are usually utilized to learn  discriminative node representation. The information aggregation can be generally formulated as follows:
\begin{equation}
    \mathbf Z^{(l+1)} = \sigma (\mathcal F(\mathbf A)\mathbf Z^{(l)} \mathbf W^{(l)} )
\end{equation}
where $\mathbf Z^{(l+1)}$ is the node embedding representation in the $l$-th layer of GNNs, $\sigma(\cdot)$ is an activation function, $\mathcal F(\mathbf A)$ denotes various graph filtering functions, and $\mathbf W^{(l)}$ is a learnable weight matrix.

\subsection{Mixup Augmentation}
Mixup was initially proposed as a data augmentation method in the field of computer vision \cite{zhang2018mixup}, where new training samples are generated  by performing linear interpolation between pairs of training instances and their corresponding labels. However, the sparsity of the graph structure and the scarcity of labeled nodes limit the scalability of Mixup when extended to graph data. To this end, several improved node mixing mechanisms suitable for graph data have been proposed to enhance GNNs \cite{wu2021graphmixup,wang2021mixup,verma2021graphmix}. These approaches not only enrich the training data but also encourage the models to learn smoother and more generalized decision boundaries. Generally, the node mixup operation can be defined by the following formula: 
\begin{equation}
\mathcal M(\mathbf a, \mathbf b) = \lambda \mathbf a + (1-\lambda) \mathbf b
\end{equation}
where $\mathbf a$ and $\mathbf b$ can be feature vectors, neighboring structure vectors, or one-hot label vectors; $\lambda$ is a mixup coefficient.

\begin{figure*}
\centering
 \includegraphics[width=13.9cm]{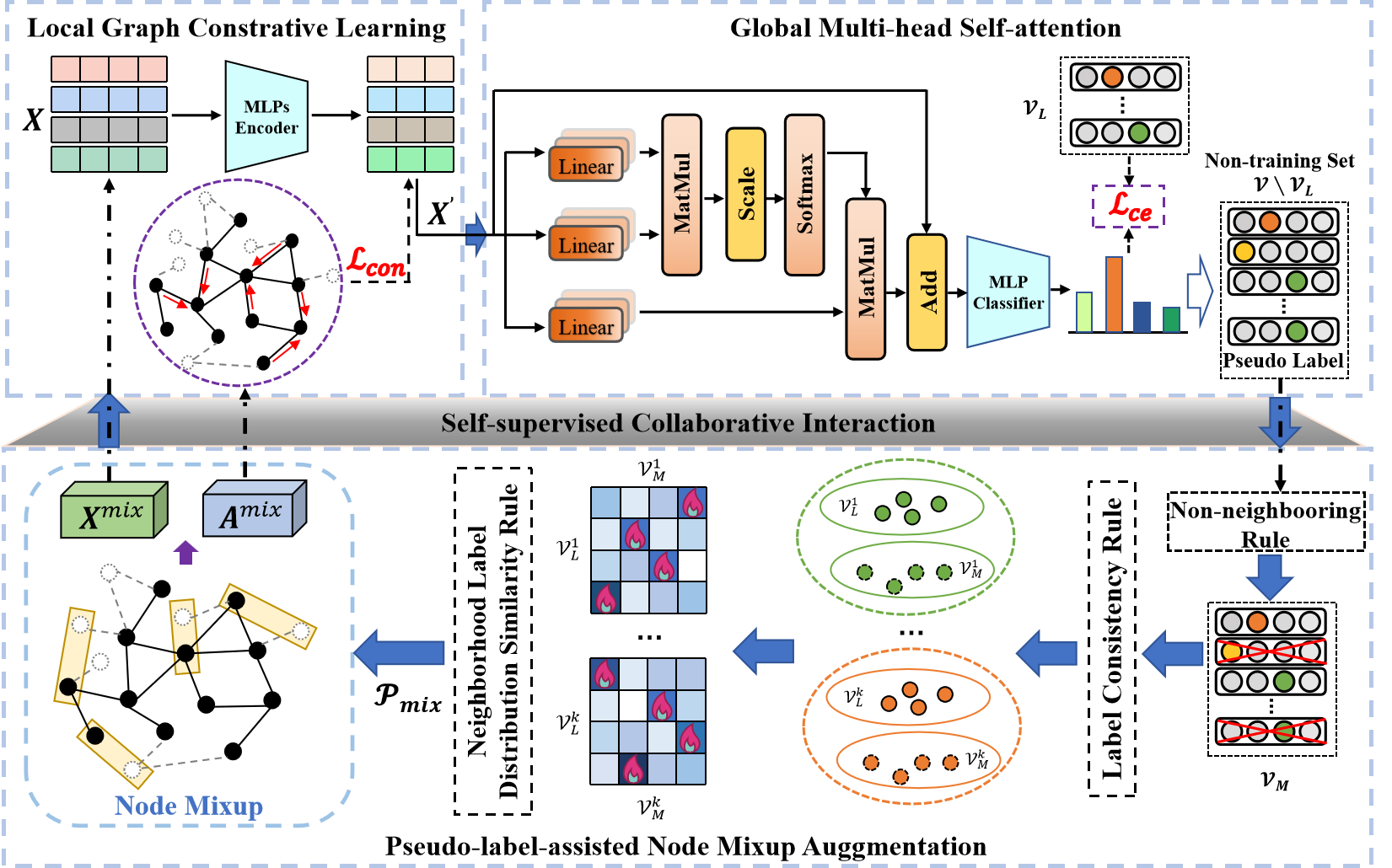}
\caption{The overall framework of the proposed ComGRL method primarily consists of the LGCL, GMSA, and PMA modules. In the first pre-training process, the LGCL and GMSA modules are unified to learn
comprehensive representation and generate global, relatively reliable pseudo labels. In the second fine-tuning process, using the global pseudo-label information, the PMA module adaptively conducts the node Mixup augmentation for the LGCL module, achieving the interaction between local and global information in a self-supervised manner.}\label{framework}
\end{figure*}

\section{Methodology}
To obtain a comprehensive representation combining local and global information, the ComGRL framework is proposed. The overall framework is shown in Figure~\ref{framework}, which mainly consists of the LGCL, GMSA, and PMA modules. Unlike most existing methods that rely on local topology and global graph structure, ComGRL integrates global and local information through discriminative representation assistance and mutual interaction. Specifically, the LGCL implicitly preserves local information between neighborhoods by the contrastive learning. Then the improved local representation is passed into the GMSA module, where multi-head self-attention explores global information in depth, and local-and-global representations are integrated. Meanwhile, a self-supervised mechanism, i.e., pseudo-label-assisted mixup augmentation, is proposed
to achieve collaboration and interaction between the local representation in LGCL and the global information in GMSA. The proposed ComGRL method is described in detail below.  

\subsection{Local Graph Contrastive Learning}
The LGCL module aims to assist the GMSA module in capturing complex neighboring topology information. Different from graph convolution in GNNs, LGCL preserves local topological relationships in an implicit manner.  
First, given the attribute features $\mathbf X $ of the graph data, a simple two-layer MLPs is utilized to obtain low-dimensional embedding representations, i.e.,
\begin{equation}
    \mathbf X' = \sigma(\sigma ( \mathbf X \mathbf W_1 ) \mathbf W_2),
\end{equation}
where $\mathbf W_1$ and $\mathbf W_2$ are the transformation parameter matrices.

Then, to improve flexibility, ComGRL implicitly preserves local information through graph contrastive learning. The $r$-hop neighbors $\mathcal N_r(v_i)$ of the target node $v_i$ are selected as positive contrastive samples, while  others nodes are treated as negative samples. A contrastive learning coefficient $\mathbf s_{ij}$ is introduced to adjust the contrastive degree, which is defined as follows
\begin{equation}\label{sc}
    s_{ij} =
\begin{cases}
    \widehat{a}_{ij} & \text{if } j \in \mathcal N_r (v_i) \\
    0 & \text{if } j \notin \mathcal N_r (v_i)
\end{cases},
\end{equation}
where $\widehat{\mathbf A} = \left(\mathbf D^{-1/2}(\mathbf A+\mathbf I) \mathbf D^{-1/2}\right)^r$ is a weighted coefficient matrix normalized by degree matrix $\mathbf D$ for multi-hop graph. And the contrastive learning loss we utilized is as follows:
\begin{equation}\label{gcl}
\mathcal L_{con} = - 1/N \sum_{i=1}^{N} \frac{\sum_{j=1}^{N} s_{ij} \mathrm {exp(sim(}\mathbf x'_i ,\mathbf x'_j )/ \tau)}{\sum_{k=1}^{N} \mathrm {exp(sim(}\mathbf x'_i ,\mathbf x'_k )/ \tau)},
\end{equation}
where $\tau$ is an adjustable temperature parameter and $\mathrm {sim(\cdot)}$ is the cosine similarity metric. By minimizing the loss \eqref{gcl} during graph contrastive learning, a larger value of $s_{ij}$ ensures that the corresponding representations of $\mathbf x'_i$ and $\mathbf x'_j$ are closer. With the multi-hop neighboring contrastive setting, the LGCL module implicitly transfers rich local information to the node representation $\mathbf X'$.

\subsection{Global Multi-head Self-attention}
The LGCL is able to learn useful embedding representation from a local perspective, while potentially overlooking important long-range dependency relationship. By incorporating structural representation, the presentation $\mathbf X'$ is integrated into the GMSA module, helping the GMSA in exploring global information more profoundly. Taking this into account, the self-attention mechanism is utilized to capture global information. The global representation learned in the $l$-th layer, $\mathbf Z^{(l+1)}$, can be obtained by the following way 
\begin{equation}\label{qkv}
\begin{aligned}
    \mathbf Z^{(l+1)} & = \text{Attention}(\mathbf Q^{(l)}, \mathbf K^{(l)}, \mathbf V^{(l)}) \\
    &= \text{softmax}\left(\frac{\mathbf{Q}^{(l)} {\mathbf{K}^{(l)}}^T}{\sqrt{d^{(l)}}} \right)\mathbf{V}^{(l)}
\end{aligned}
\end{equation}
where 
\begin{equation}
    \mathbf Q^{(l)} = \mathbf Z^{(l)}\mathbf W_Q^{(l)},
    \mathbf K^{(l)} = \mathbf Z^{(l)}\mathbf W_K^{(l)},
    \mathbf V^{(l)} = \mathbf Z^{(l)}\mathbf W_V^{(l)}
\end{equation}
in which $\mathbf Q^{(l)}$, $\mathbf K^{(l)}$, and $\mathbf V^{(l)}$ are the representation of the query, keys, and values, respectively. The $d^{(l)}$ is a scaling parameter, and the $\mathbf W_Q^{(l)}$, $\mathbf W_K^{(l)}$, and $\mathbf W_V^{(l)}$ are learnable parameter matrices, which help reduce redundancy in self-attention.  Besides,
we denote initial $\mathbf Z^{(0)}$ as the local representation $\mathbf X^{'}$. In Eq.~\eqref{qkv}, the self-attention mechanism calculates the global structural similarity for any node pair, thus aggregating information from the global perspective. Further, to capture rich and diverse global interactions, the self-attention could be extended to multi-head. i.e.,
\begin{equation}\label{ms}
    \mathbf Z^{(l+1)} = \text{Concat} (\mathrm {head}_1^{(l+1)}, \cdots, \mathrm {head}_h^{(l+1)}) \mathbf W_{h}^{(l+1)}  
\end{equation}
where $\mathrm {head}_i^{(l+1)} = \text{Attention} (\mathbf Q_i^{(l)},\mathbf K_i^{(l)}, \mathbf V_i^{(l)})$ is representation calculated by Eq.~\eqref{qkv} for $i$-th head, $\text{Concat} (\cdot) $ represents the concatenation of various representations, and $\mathbf W_{h}^{(l+1)}$ is a learnable parameter matrix. Each $\mathrm {head}_i$ explores specific global semantic information in the head-specific interaction space $\{\mathbf{Q}_i^{(l)}, \mathbf{K}_i^{(l)}, \mathbf{V}_i^{(l)}\}$, which is then combined to create a richer and more comprehensive representation.

Due to the high computational complexity of the standard self-attention mechanism in Eq.~\eqref{qkv}, which is $\mathcal O(N^2)$, an efficient version \cite{shen2021efficient} with the complexity of $\mathcal O(N)$ is utilized to calculate the attention coefficient, i.e.,
\begin{equation}
\begin{aligned}
&\text{Attention}(\mathbf Q^{(l)}, \mathbf K^{(l)}, \mathbf V^{(l)})\\=& \sigma_1(\mathbf Z^{(l)}  \mathbf W_Q^{(l)})(\sigma_2(\mathbf Z^{(l)}  \mathbf W_K^{(l)} )^T \mathbf Z^{(l)}  \mathbf W_V^{(l)})
\end{aligned}
\end{equation}
where both $\sigma_1$ and $\sigma_2$ are the softmax functions for normalization. Then, with multi-head representation $\mathbf Z^{(l+1)}$ in Eq.~\eqref{ms}, to obtain a more powerful and complete representation, the representations across heads and layers are simultaneously combined as follows \cite{Vaswani2017attention}:
\begin{equation}
\begin{aligned}
\mathbf Z^{(l+1)} & = \mathbf Z^{(l+1)} + \mathbf Z^{(l)}  \\
\mathbf Z^{(l+1)}& =\mathbf W_{3} \ \phi(\mathbf W_4 \text{LN}(\mathbf Z^{(l+1)})) + \mathbf Z^{(l+1)} \\
\end{aligned}
\end{equation}
where $\phi$ represents activation function such as ReLU, and $\text{LN}(\cdot)$ is the layer normalization operation. The cross-layer concatenation allows the GMSA module to leverage different valuable and richer contextual information. By integrating local representation with self-attention, the GMSA can mitigate the influence of feature redundancy and explore more accurate global information.  

Consequently, the final representation $\mathbf Z$ from multi-head self-attention is utilized to train the network and classifier with the cross-entropy loss:
\begin{equation}
\mathcal L_{ce} = - \sum_{v_i \in \mathcal V_L}\sum_{c=1}^{k} y_{ic} \text{log} \ z_{ic}
\end{equation}

\subsection{Pseudo-label-assisted Mixup Augmentation}

Now, the LGCL module has been naturally connected with the GMSA module in the network architecture. Nevertheless, a deeper connection between them, particularly in terms of mutual collaboration, has not yet been designed. Basically, LGCL is primarily utilized for local representation learning, which is sensitive to graph structure. However, practical graphs are often sparse and noisy, which can negatively affect local message-passing. Fortunately, after several epochs of network training with local-and-global information propagation in a semi-supervised setting, additional and relatively reliable supervision information from unlabeled data can be obtained through the pseudo-label alignment strategy. Inspired by this, this study proposes a self-supervised PMA module, which unifies the LGCL and GMSA modules in a collaborative way. It effectively mixes labeled and unlabeled data to achieve information augmentation (including attributes and structure) for the LGCL module, guided by the pseudo-label information from the GMSA module. For the PMA module, reliable node mixup pairs constructed from labeled and unlabeled nodes are essential for facilitating direct communications between them. This effectively expands the receptive field of message-passing in local representation learning, improving both its robustness and discriminability‌ for global representation learning. Guided by high confidence pseudo labels obtained after pre-training, a triple sampling strategy is developed to construct node mixup pairs. This strategy primarily adheres to the following principles: non-neighboring nodes, label consistency, and similarity in neighborhood label distribution.

\textbf{Non-neighboring.} Considering the limited label information, unlabeled data is usually mixed with labeled data to enhance the role of unlabeled data in the training process. To further propagate supervisory information to the unlabeled nodes in the graph and expand the receptive field of message-passing, unlabeled nodes that are beyond the $r$-hop neighbors of all labeled nodes are considered as candidates for node mixup augmentation. Let $\mathcal V_L$ denote the set of the labeled nodes, candidate nodes for Mixup augmentation are selected as follows
\begin{equation}
\mathcal{V}_M = \left\{v_j \in \mathcal V \backslash \mathcal V_L \mid \forall v_i \in \mathcal{V}_L, v_j \notin \mathcal{N}_r(v_i)\right\}
\end{equation}
\textbf{Remark: Although we select unlabeled nodes that are not close to any labeled nodes (beyond $r$-hops), due to the sparse labeling and connection between nodes, there are still sufficient candidate nodes in $\mathcal V_M$.} Selecting nodes that are far away from the labeled nodes to mix structure with labeled nodes also can to some extent avoid excessive reliance on fixed local structure in graph contrastive learning due to adaptive structural augmentation.

\textbf{Label Consistency.} After pre-training for several epochs, the predicted assignment probabilities become relatively reliable and
serve as pseudo labels for the unlabeled nodes, acting as true labels in the node mixup pairs construction. To enhance the reliability of Mixup, the label consistency principle is important.  Suppose the labels of $v_i \in \mathcal V_L$ and $v_j \in \mathcal V_M$ are $\mathbf y_i$ and $\hat{\mathbf{y}}_j$, where $\mathbf y_i$ is the true label and $\hat{\mathbf{y}}_j$ is the pseudo label. Then, node pairs for Mixup augmentation are constructed class by class, i.e., $(\mathcal{V}_L^c, \mathcal{V}_M^c)$ for each class $c \in \{ 1, 2, \dots, k\}$, where
\begin{equation}
\begin{aligned}
\mathcal{V}_L^c &= \{ v_i \in \mathcal{V}_L \mid \mathbf y_i = \mathbf 1_c \},  \\
\mathcal{V}_M^c &= \{ v_j \in \mathcal{V}_M \mid \hat{\mathbf y}_j = \mathbf 1_c \}
\end{aligned}
\end{equation}
where $\mathbf 1_c$ is a one-hot vector corresponding to class $c$.

\textbf{Neighborhood Label Distribution Similarity.} In a graph, the characteristics of a node are also influenced by its neighbors. Specifically, nodes with similar neighbor distributions are beneficial for improving the effect of mixup augmentation. The graph constructed by the pseudo labels should be more accurate than the original graph. Hence, the Neighborhood Label Distribution (NLD) \cite{zhu2023heterophily} is utilized to construct the graph $\mathbf S$ class by class, in which the node pair with the highest similarity in neighbor distributions from sets $\mathcal{V}_L^c$ and $\mathcal{V}_M^c$ is finally selected for Mixup augmentation. The neighborhood label distribution of any node $v_i$ in $\mathcal{V}_L^c \cup \mathcal{V}_M^c$ is defined as:
\begin{equation}
\mathbf{f}_i = \frac{1}{|\mathcal{N}(v_i)|} \sum_{v_j \in \mathcal{N}(v_i)} \overline{\mathbf{y}}_j
\end{equation}
where $\overline{\mathbf{y}}_j = \mathbf y_j$ is true label if $v_j \in \mathcal V_L$, and $\overline{\mathbf{y}}_j = \hat{\mathbf y}_j$ is pseudo label if $v_j \in \mathcal V \backslash \mathcal V_L$.

With the neighborhood label distribution $\mathbf F$, we try to
obtain a more accurate similarity graph by improving the high confidence assignments. Specifically,  the sharpening trick \cite{berthelot2019mixmatch}, i.e., exponentiating and normalizing each distribution in $\mathbf F$, is utilized to more higher confidence:
\begin{equation}
\mathbf{f}_i^{'} =\frac {\mathbf{f}_i^{\frac{1}{\beta}}}{\sum_{j=1}^{k} f_{ij}^{\frac{1}{\beta}}}
\end{equation}
where $0 < \beta \leq 1$ is a temperature parameter that adjusts the sharpness of target neighborhood label distribution. The smaller the value of $\beta$,  the sharper the distribution. Finally, the target distribution $\mathbf{F}^{'}$ can help obtain a more accurate and compact neighborhood distribution similarity matrix $\mathbf S \in [0,1]^{|\mathcal{V}_L^c|\times |\mathcal{V}_M^c|}$ for node pairs $(\mathcal{V}_L^c, \mathcal{V}_M^c)$ with a class by class way, i.e., 
\begin{equation}
s_{ij}= \frac{\mathbf{f'}_i \cdot \mathbf{f'}_j}{\|\mathbf{f'}_i\|_2  \|\mathbf{f'}_j\|_2}
\end{equation}
The node pair with most similar neighborhood label distribution is selected as candidate set for Mixup augmentation, i.e., $\mathcal P_{mix}=\{(v_i, v_j)\mid v_i \in \mathcal{V}_L^c, v_j \in \mathcal{V}_M^c, v_j = \mathop{\arg\max}\limits_{v_l \in \mathcal{V}_M^c} s_{il}^c, \forall c \in \{1, 2, \cdots, k\}\}$. This triple sampling strategy preserves nodes with longer distance but consistent class information, resulting in a reasonable and reliable mechanism for the Mixup augmentation.

The attributes and structure are all crucial for message-passing in graph representation learning. Thus, the attribute feature and topological structure are mixed to generate new labeled data and graph structure as follows
\begin{equation}
\begin{aligned}
\mathcal{D}_M = \{ (\mathbf {\tilde{x}}_i,\mathbf {y}_i)  \mid 
 \mathbf {\tilde{x}}_i =\mathcal M(\mathbf {x}_i, \mathbf {x}_j), (v_i, v_j) \in \mathcal P_{mix}\}
 \end{aligned}
\end{equation}
and
\begin{equation}
\begin{aligned}
\mathbf {A}_{i,:}^{mix} & = \mathcal M(\mathbf {A}_{i,:}, \mathbf {A}_{j,:}), (v_i, v_j) \in \mathcal P_{mix}  \\
\mathbf {A}_{:,i}^{mix} & = \mathcal M(\mathbf {A}_{:,i}, \mathbf {A}_{:,j}), (v_i, v_j) \in \mathcal P_{mix}
\end{aligned}
\end{equation}
where $\mathbf A_{i,:}$ and $\mathbf A_{:,i}$ are the $i$-th row and column vector of original graph adjacency matrix $\mathbf A$. In the subsequent training, the attribute and structure of mixed nodes are dynamically updated and used as input of the LGCL module, automatically concentrating the LGCL and GMSA modules in a self-supervised way. In addition, the structure augmentation could alleviate the negative impact of graph noise for node classification.

\subsection{Training Process}
A two-stage training strategy is used, where the first stage involves a pre-training process to generate relatively reliable pseudo labels, and the second stage focuses on a self-supervised fine-tuning process using pseudo-label-assisted node mixup augmentation. The comprehensive loss function utilized in training is defined as follows:
\begin{equation}\label{loss_f}
\mathcal L = \mathcal L_{ce} + \alpha\mathcal L_{con}
\end{equation}
where $\alpha$ is a non-negative balanced parameter. In the first pre-training stage, using the  original attribute features and graph structure, the LGCL and GMSA modules are unified to learn a comprehensive node embedding representation. More importantly, after the warm-up epochs of training, the higher-confidence pseudo labels are obtained to provide prior guidance for node Mixup augmentation. In the second fine-tuning stage, pseudo-label-assisted node Mixup augmentation achieves node feature augmentation and graph structure rectification for subsequent local representation learning, implicitly enlarging the receptive fields of message-passing. The pseudo-label-assisted node Mixup augmentation   helps the LGCL module learn a more effective and discriminative representation for GMSA. The training process in the second stage can be regarded as a self-supervised feedback process, because pseudo labels are obtained by unifying the LGCL and GMSA modules, and the pseudo-label information subsequently guides the node augmentation for the LGCL in turn. Additionally,  the PMA module is capable of achieving the interaction between local and global information.

\section{Experiments}

\subsection{Datasets}
To comprehensively evaluate the effectiveness the proposed ComGRL on semi-supervised node classification tasks, six widely used datasets are utilized in this experiment. These include four classical citation network datasets including Cora, Citeseer, Pubmed \cite{yang2016revisiting}, and CoraFull \cite{chen2023dual}, where CoraFull is an expanded version of the Cora dataset with more samples and classes. Additionally, two coauthor
networks CS and Physics datasets \cite{shchur2018pitfalls} are also used. Table~\ref{dataset} summarizes the detailed statistics and Train/Val/Test splits of six datasets. 

\begin{table*}
    \centering
       \renewcommand\arraystretch{0.95}
        \begin{tabular}{cccccccccccccc}
           \hline
           & \multicolumn{5}{c}{Statistics} & &\multicolumn{7}{c}{Parameter Settings} \\
            Datasets  & \cellcolor{lightgray}{Nodes} & \cellcolor{lightgray}{Edges} & \cellcolor{lightgray}{Features} & \cellcolor{lightgray}{Class} & \cellcolor{lightgray}{Train/Val/Test} & &\cellcolor{lightgray1}{$\alpha$} & \cellcolor{lightgray1}{$\tau$} & \cellcolor{lightgray1}{$r$} & \cellcolor{lightgray1}{$T_{pre}$} & \cellcolor{lightgray1}{$T_{total}$} & \cellcolor{lightgray1}{$lr$} & \cellcolor{lightgray1}{$dr$} \\
            \hline
            Cora     & 2708 & 10556 & 1433 & 7 & 140/500/1000 &  & 1.0 &1.8 & 4 & 300 & 500 & 5e-4 & 0.4 \\
            Citeseer&   3327& 9104 & 3703 & 6  & 120/500/1000 & & 0.1 & 0.7 & 4  & 100  & 1000 & 3e-4  & 0.4\\
            Pubmed  & 19717 & 88648 & 500 & 3 & 60/500/1000 & & 1.0 & 1.0 & 4 & 300 & 1000 & 4e-4 & 0.4 \\
            CS  & 18333 & 163788 & 6805  & 15  & 300/450/17583 & & 1.0 & 1.8 & 4  & 20  & 600  & 2e-4 & 0.1 \\
            Physics  & 34493 & 495924 & 8415  & 5  & 100/150/34243 & & 2.0 & 1.0 & 4  & 300 &1000 & 4e-4  &  0.1 \\
            CoraFull & 19793 & 126842 & 8710  & 70  & 1400/500/1000 & & 1.0 & 2.2 & 3  & 40 & 400 & 4e-4 & 0.4\\
            \hline
        \end{tabular}
    \caption{The statistics of all used datasets and parameter settings of proposed ComGRL method.}\label{dataset}
\end{table*}

\begin{table*}
    \centering
    \begin{tabular}{lcccccc}
        \hline
        Datasets  & Cora & Citeseer & Pubmed & CS & Physics  & CoraFull \\
        \hline
        MLP & $61.60 \pm 0.60$ &  $61.00 \pm 1.00$ & $74.20 \pm 0.70$& $88.30 \pm 0.70$	& $88.90 \pm 1.10$	&$5.10\pm0.30$  \\
        DeepWalk  & $70.70 \pm 0.60$& $51.40\pm 0.50$	& $76.80\pm 0.60$& 	$84.60\pm 0.20$& $90.62\pm1.28$  & $48.40\pm 0.60$ \\
        Chebyshev &	$81.40 \pm 0.70$ &	$70.20\pm 1.00$ &	$78.40 \pm 0.40$	& $91.50\pm 0.00$	& $91.49\pm0.73$	& $53.40\pm 0.50$  \\
        GCN	& $81.50\pm 0.50$ &  $70.30\pm 0.70$ &  $79.00\pm 0.70$  & $91.10\pm 0.50$  &	$92.80\pm 1.00$  & $56.70\pm 0.40$ \\
        GAT &  $83.00\pm 0.70$  &	$72.50\pm 0.70$  & $79.00\pm 0.30$  & $90.50 \pm 0.60$  & $92.50 \pm 0.90$ &	$57.10\pm 1.00$ \\
        SGC &	$81.00 \pm 0.00$ &	$71.90 \pm 0.10$ &	$78.90\pm 0.00$ &	$91.00\pm 0.00$ &$92.20\pm 0.60$	&  $58.30 \pm 1.20$  \\
        APPNP &  $83.30 \pm 0.50$ &  $71.80 \pm 0.50$ &  $80.10\pm 0.20$  & $72.10\pm 0.70$  &	$93.20\pm 0.40$  &   $57.60\pm 0.40$  \\
        UPS  &	$82.17\pm 0.50$ &	$72.97\pm 0.70$ &	$78.56\pm 0.90$ &  $91.26\pm 0.40$  &	$92.45\pm 1.10$  & $57.98 \pm 1.04$  \\
        HOG-GCN	& $81.20\pm 0.30$ & $69.20\pm0.60$  &  $78.80\pm1.20$	& 
        $90.70\pm0.40 $  &  $92.80\pm0.30 $   &	$48.40\pm0.70 $  \\
        LGD-GCN	& $\underline{85.20\pm 0.60}$ & $\underline{74.40\pm0.50}$  &  $\underline{81.50\pm0.60}$	& 
         $92.48\pm0.82$  &  $93.14\pm0.66$  & $\underline{58.38\pm0.27}$  \\
          RNCGLN	& $84.46\pm 0.35$ & $73.96\pm0.30$  &  $81.16\pm0.64$	& 
        $\underline{92.69\pm0.41} $  &  $92.78\pm0.86 $   &	$57.08\pm 1.24 $  \\
        
        Graphmix  &  $83.32\pm 0.18$ & 	$73.08\pm 0.23$ &  $81.10\pm 0.78$  & 	$90.57\pm1.00$	&  $92.90\pm 0.40$ &   $57.66 \pm 0.53$   \\
        Nodemixup &	$83.52\pm 0.30$ &	$74.30\pm 0.10$  &   $81.26\pm 0.30$ & $92.69\pm 0.20$ &	$\underline{93.87\pm 0.30}$ &	$58.04 \pm 1.29$  \\
        \hline
        ComGRL & $\mathbf{85.39\pm 0.39}$ &	$\mathbf{75.14\pm0.44}$ &	$\mathbf{82.30\pm 0.60}$ &	$\mathbf{93.58\pm 0.36}$  &  $\mathbf{94.08\pm 0.17}$  &	$\mathbf{59.54\pm 0.11}$   \\
        \hline
    \end{tabular}
    \caption{The node classification results on the all used datasets.}
    \label{tabresult}
\end{table*}

\subsection{Experimental Settings}
There are three crucial hyper-parameters, including the balanced coefficient $\alpha$ in Eq.~\eqref{loss_f} of the loss function, the temperature parameter $\tau$ in Eq.~\eqref{gcl} of the contrastive loss, and the number of neighbors $r$ in Eq.~\eqref{sc}. The detailed parameter settings are shown in Table~\ref{dataset}. For the network architecture, we uniformly set the number of heads in the multi-head self-attention mechanism to 8 and employ a 2-layer attention network. The dimensionality of the hidden layers is consistent with that of the MLPs hidden layers, both set to 128, and we utilize the LeakyReLU activation function. The model is run for $T_{total}$ epochs in total, where the first $T_{pre}$ epochs are for the pre-training stage to obtain reliable pseudo labels.

\subsection{Baselines}
To achieve a thorough evaluation of the proposed ComGRL, several representative methods are selected as baselines for comparison, including MLP \cite{taud2018multilayer}, DeepWalk \cite{perozzi2014deepwalk}, Chebyshev \cite{defferrard2016convolutional}, GCN \cite{kipf2017semi}, GAT \cite{velickovic2017graph}, SGC \cite{wu2019simplifying}, APPNP \cite{gasteiger2018predict}, UPS \cite{rizve2021defense}, HOG-GCN\cite{wang2022powerful}, LGD-GCN\cite{guo2022learning}, 
RNCGLN \cite{zhu2024robust}, Graphmix \cite{verma2021graphmix} and Nodemixup\cite{lu2024nodemixup}.

\subsection{Experimental Results and Analysis}
The node classification results of the proposed ComGraph and all compared methods are presented in Table~\ref{tabresult}, where the best results are highlighted with bold and the sub-optimal ones are underscored with underline. As shown in Table~\ref{tabresult}, some crucial observations can be concluded as follows: 
\begin{itemize}
    \item \textbf{The proposed ComGRL achieves optimal node classification performance across the six datasets.} In particular, ComGRL achieves significant improvements of $0.19\%$, $0.74\%$, $0.80\%$, $0.89\%$ $0.21\%$, and $1.16\%$ over the second-best competitors on all datasets, respectively. This success is primarily due to ComGRL's effective integration of local information into global representations, combined with   the pseudo-label-assisted node Mixup augmentation module.  This module facilitates the interaction between LGCL and GMSA, allowing them to mutually enhance one another and achieve a comprehensive embedding representation for node classification. 
    \item Node classification results of Mixup augmentation based methods (Graphmix and Nodemixup) are generally superior than classical GNN methods (GCN and GAT). The reason is that node mixup augmentation can enhance the attribute and structure, thereby promoting message-passing and thus improving performance. However, ComGRL surpasses these Mixup augmentation based baselines.  This success is attributed to its ability to explore both global and local information effectively. More importantly, ComGRL adaptively updates the node mixup information under self-supervision with high-confidence pseudo-labels.
    \item Compared with other GNNs methods specifically designed for integrating global and local information (i.e. LGD-GCN, HOG-GCN, and RNCGLN), the average classification performance of ComGRL improves by $0.19\%$-$5.94\%$. This further demonstrates that integrating global and local information through representation delivery and achieving collaborative interaction between them are beneficial to obtain comprehensive embedding representation.
\end{itemize}
\subsection{Ablation Analysis}
The ablation experiments are designed to evaluate the effectiveness of ComGRL from two aspects. The first ablation study progressively removes crucial components (including LGCL, GMSA, and PMA) from complete ComGRL and analyzes the performance of the degraded models. From the results shown in Table~\ref{abl1}, the performance of ComGRL drops when the above components are
removed, demonstrating that integrating both global and local information with node augmentation can prominently enhance the discriminative ability of node representation. Specifically, the performance of ComGRL without the LGCL module dramatically declines, which indicates that, without exploring local information, GMSA cannot accurately and profoundly capture global correlations.
\begin{table}
    \centering
    \renewcommand\tabcolsep{2.0pt}
    \resizebox{0.49\textwidth}{!}
    {
    \begin{tabular}{ccccccccc}
        \hline
        LGCL  & GMSA & PMA & Cora & CiteSeer & Pubmed & CS & Physics & CoraFull \\
        \hline
        \xmark  & \cmark & \cmark & $61.60$ & $57.70$ & $71.20$ & $89.59$ & $88.49$ & $42.66$ \\
        \cmark & \xmark & \cmark & $80.40$ & $72.20$ & $80.20$ & $90.94$ & $90.78$ & $57.34$  \\
        \cmark & \cmark & \xmark & 83.55 & $72.56$ & $79.80$ &  $92.40$ & $92.45$ & $57.13$\\
         \cmark & \cmark & \cmark & $\mathbf{85.39}$ & $\mathbf{75.14}$ & $\mathbf{82.30}$ & $\mathbf{93.58}$ & $\mathbf{94.08}$ & $\mathbf{59.54}$ \\
        \hline
    \end{tabular}}
    \caption{Comparisons of ComGRL with its ablation variants without LGCL, GMSA, and PMA modules, respectively.}\label{abl1}
\end{table}

\begin{figure}
\centering
\subfigure{
\includegraphics[width=4cm]{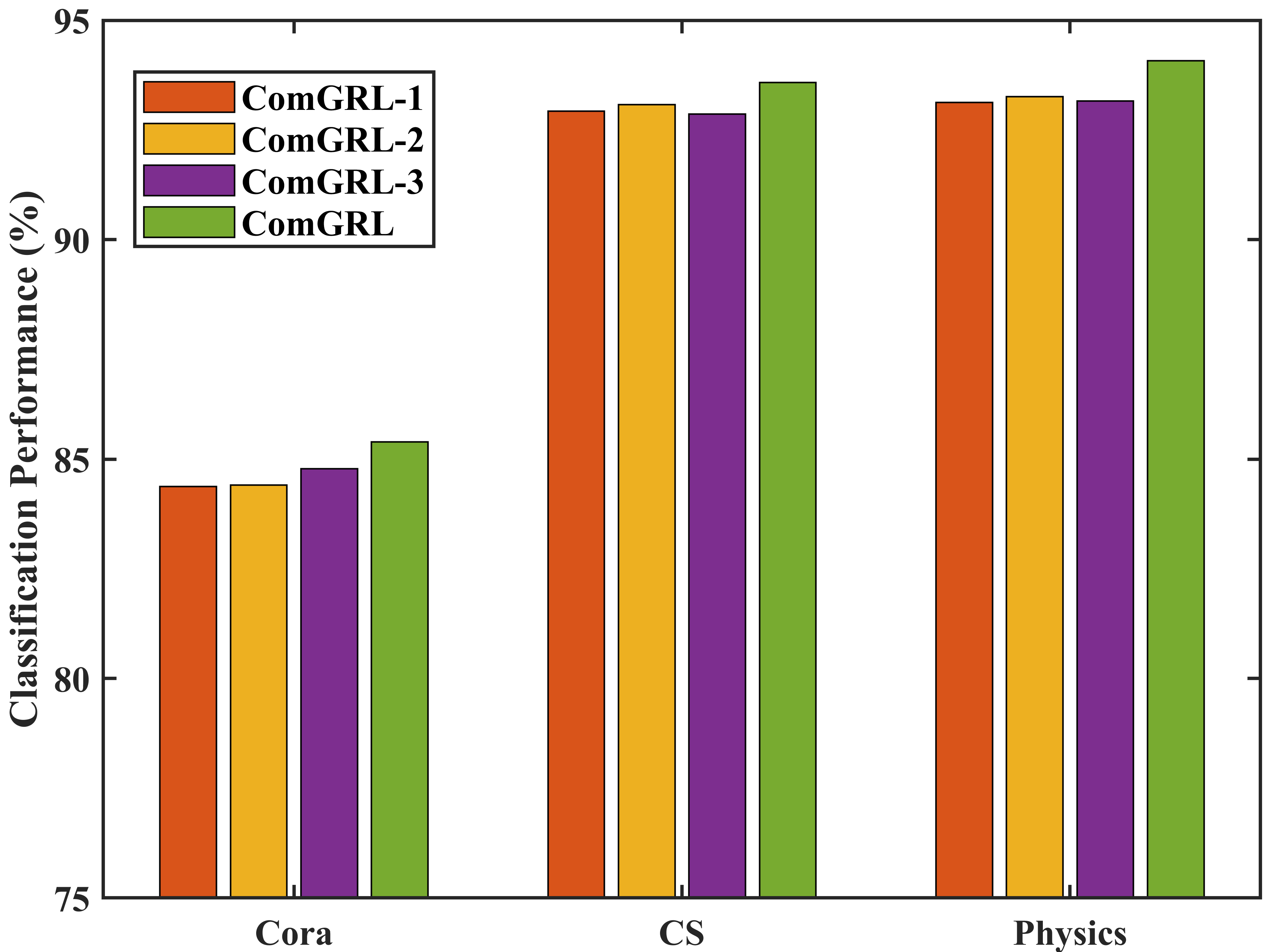}
}
\hspace{-2mm}
\subfigure{
\includegraphics[width=4cm]{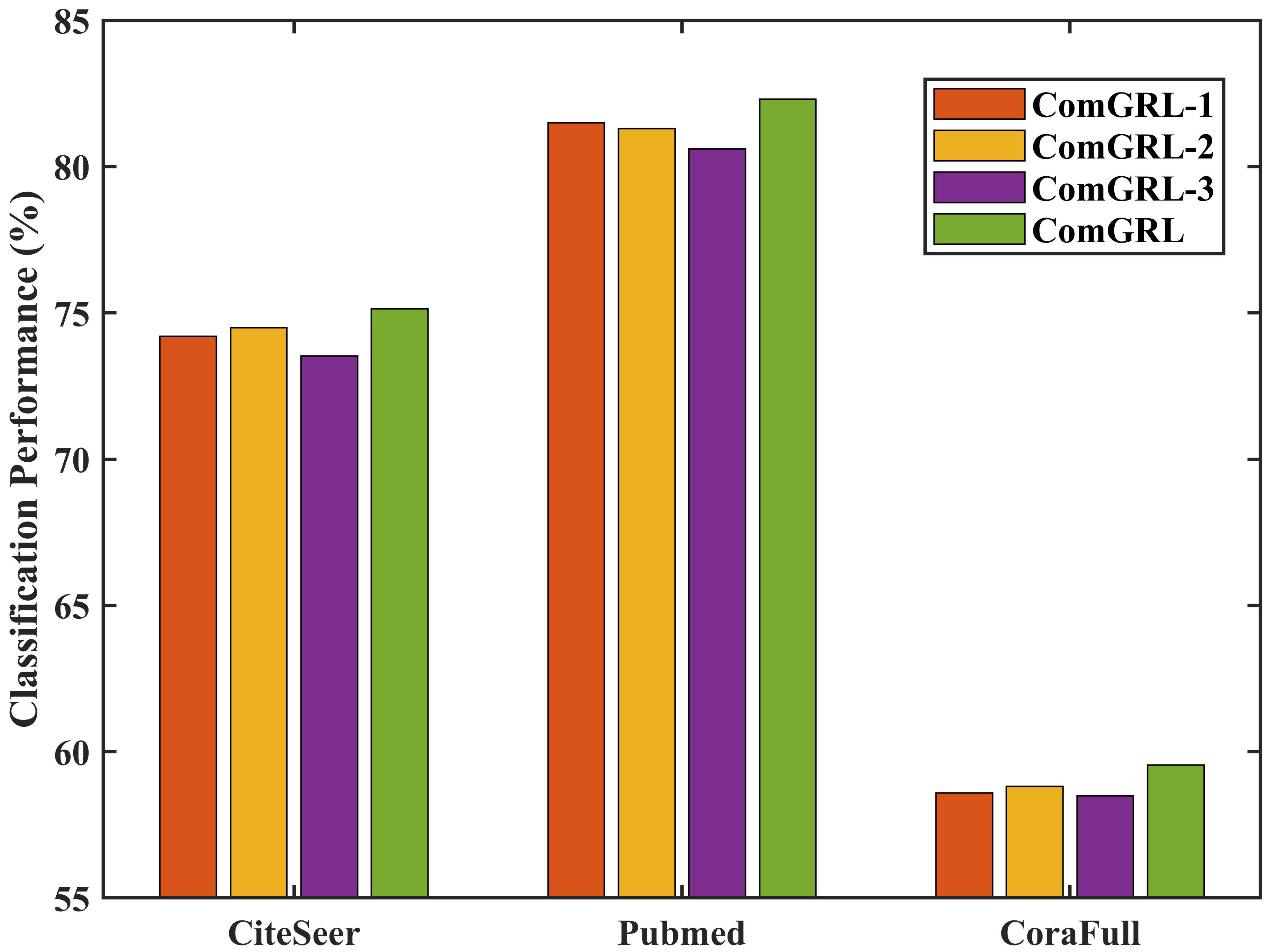}
}
\caption{The experimental results across different ablation settings in PMA module.}\label{abl2}
\end{figure}

The second ablation experiment assesses the influence of the three node pair construction principles in PMA. The variants are as follows: 1) ComGRL-1: removing the non-neighboring constraint between unlabeled and labeled nodes; 2) ComGRL-2: removing the label consistency principle; 3) ComGRL-3: removing the neighborhood label distribution similarity principle. From the ablation results in Figure~\ref{abl2}, all three variants perform  worse than complete ComGRL, which indicates that the above three node pair construction principles are all necessary and have a positive impact on node self-supervised augmentation. These principles  consciously expand the receptive field of message-passing. It is noteworthy that, in most cases, the performance degradation of ComGRL-3 is the  most significant. The reason is that NLD consistency improves the probability of mix between nodes with similar neighbor distributions, thereby
enhancing the confidence of information augmentation and generalization.

\subsection{Robustness Analysis}
In the fine-tuning training, the PMA module can rectify the graph structure, and the self-supervised augmentation strategy helps alleviate the training dependence on label information. We investigate the robustness of ComGRL to graph noise and label noise. Three classic node classification methods  dealing with noises are selected as competitors, including NRGNN \cite{dai2021nrgnn}, LAFAK \cite{zhang2020adversarial}, and RNCGLN. The experimental results on the Cora and CiteSeer datasets under various noise ratios are shown in Table~\ref{tabnosie}, i.e., $\{lnr, gnr\} = \{0.3, 0.1\}$ and $\{lnr, gnr\} = \{0.1, 0.3\}$, where $lnr$ and $gnr$ are ratios of label noise and graph noise, respectively. According to Table~\ref{tabnosie}, we  observe that ComGRL achieves optimal or suboptimal performance under different proportions of graph noise and label noise. This demonstrates that ComGRL can obtain more powerful and robust embedding representations. This superiority is mainly attributed to the local-and-global collaborative learning and self-supervised node augmentation, which helps alleviate the negative impact of noises.  

\begin{table}
    \centering
        \renewcommand\tabcolsep{2.0pt}
   \resizebox{0.49\textwidth}{!}{
    \begin{tabular}{c cccc}
        \hline
          & \multicolumn{2}{c}{Cora} & \multicolumn{2}{c}{Citeseer} \\
         $\{lnr, gnr\}$    & \cellcolor{lightgray}{\{0.3, 0.1\}}    & \cellcolor{lightgray}{ \{0.1, 0.3\}}  & \cellcolor{lightgray1}{\{0.3, 0.1\} }    & \cellcolor{lightgray1}{\{0.1, 0.3\}}  \\
        \hline
         NRGNN   & $70.90\pm3.88$   & $72.40\pm1.98$  & $66.80\pm2.31$   & $64.46\pm3.23$  \\
         LAFAK   & $79.34\pm0.54$    & $74.34\pm1.34$  & $69.14\pm1.28 $   & $66.42\pm1.04 $  \\
         RNCGLN   & $\mathbf{81.73\pm0.33}$   & $80.60\pm1.06$  & $71.83\pm0.33 $   & $70.20\pm1.41$  \\
         ComGRL   & $80.92\pm0.07$    & $\mathbf{81.17\pm0.72}$  & $\mathbf{72.27\pm0.35}$   & $\mathbf{71.13\pm0.74}$  \\
        \hline
    \end{tabular}}
    \caption{The results of node classification under different ratios of graph noise and label noise.}
    \label{tabnosie}
\end{table}

\section{Conclusion}
In this paper, a novel comprehensive graph representation learning framework was proposed to adaptively integrate local and global information, where the local information and global correlation are simultaneously explored by graph contrastive learning and multi-head self-attention strategies. The local representation from LGCL was delivered to GMSA, which is capable of capturing the most discriminative global correlations and effectively integrating their complementarity for comprehensive representation learning. With the global pseudo-label information, the PMA module with a triple sampling strategy achieved node augmentation for graph contrastive learning, expanding the receptive field and facilitating deep collaboration between the LGCL and GMSA modules. The experimental results conducted on six datasets verified the effectiveness
of ComGRL for node classification tasks.

\bibliographystyle{named}
\bibliography{ref}

\end{document}